\documentclass[10pt,twocolumn,letterpaper,usenames,dvipsnames]{article}

\usepackage{cvpr}
\usepackage{float}
\usepackage{times}
\usepackage{epsfig}
\usepackage{graphicx}
\usepackage{amsmath}
\usepackage{amssymb}
\usepackage{subcaption}
\usepackage{tikz}
\usepackage{xcolor,colortbl}
\usepackage[]{algorithm2e}
\usepackage{csvsimple}
\usepackage{booktabs}
\usepackage[export]{adjustbox}

\newcommand{\comment}[1]{}

\definecolor{Gray}{gray}{0.85}
\newcolumntype{g}{>{\columncolor{Gray}}c}

\definecolor{floor}{RGB}{0,128,0}
\definecolor{furniture}{RGB}{0,128,128}
\definecolor{structure}{RGB}{0,0,255}
\definecolor{props}{RGB}{200,200,0}

\definecolor{oil_bottle}{RGB}{0,0,255}
\definecolor{tissue_box}{RGB}{255,0,0}
\definecolor{blue_funnel}{RGB}{153,153,0}
\definecolor{drill}{RGB}{153,200,200}
\definecolor{cracker_box}{RGB}{102,0,52}
\definecolor{tomato_soup}{RGB}{0,102,102}
\definecolor{spam}{RGB}{0,0,0}

\usepackage[pagebackref=true,breaklinks=true,letterpaper=true,colorlinks,bookmarks=false]{hyperref}

\cvprfinalcopy 

\def\cvprPaperID{6206} 

\ifcvprfinal\fi
\begin{document}


\title{S4-Net: Geometry-Consistent Semi-Supervised Semantic Segmentation}


\author{Sinisa Stekovic$^1$\\
\and
Friedrich Fraundorfer$^1$\\
\and
Vincent Lepetit$^{1,2}$\\
\and
$^1$Institute for Computer Graphics and Vision, Graz University of Technology, Graz, Austria \\
$^2$Laboratoire Bordelais de Recherche en Informatique, University of Bordeaux,
Bordeaux, France \\
{\tt\small \{sinisa.stekovic, fraundorfer, lepetit\}@icg.tugraz.at}
}

\maketitle

\begin{abstract}

We show  that it is  possible to learn  semantic segmentation from  very limited
amounts of  manual annotations,  by enforcing  geometric 3D  constraints between
multiple views. More exactly, image locations corresponding to the same physical
3D  point  should all  have  the  same label.   We  show  that introducing  such
constraints during  learning is  very effective,  even when  no manual  label is
available for a  3D point, and can  be done simply by  employing techniques from
'general' semi-supervised learning to the  context of semantic segmentation.  To
demonstrate  this idea,  we use  RGB-D image  sequences of  rigid scenes,  for a
4-class segmentation  problem derived from  the ScanNet dataset.   Starting from
RGB-D sequences  with a few  annotated frames, we  show that we  can incorporate
RGB-D sequences without any manual annotations to improve the performance, which
makes our  approach very convenient.   Furthermore, we demonstrate  our approach
for semantic segmentation  of objects on the LabelFusion dataset,  where we show
that one manually labeled image in a scene is sufficient for high performance on
the whole scene.

\end{abstract}



\section{Introduction}

\begin{figure}[t]
	\centering
	\begin{subfigure}[b]{1.\linewidth}
		\begin{tikzpicture}
		
		\node[label={[label distance=-2mm]\textit{\small $I_1$}}] (simg) at (-8.0, 0.0) {\includegraphics[width=0.2\linewidth]{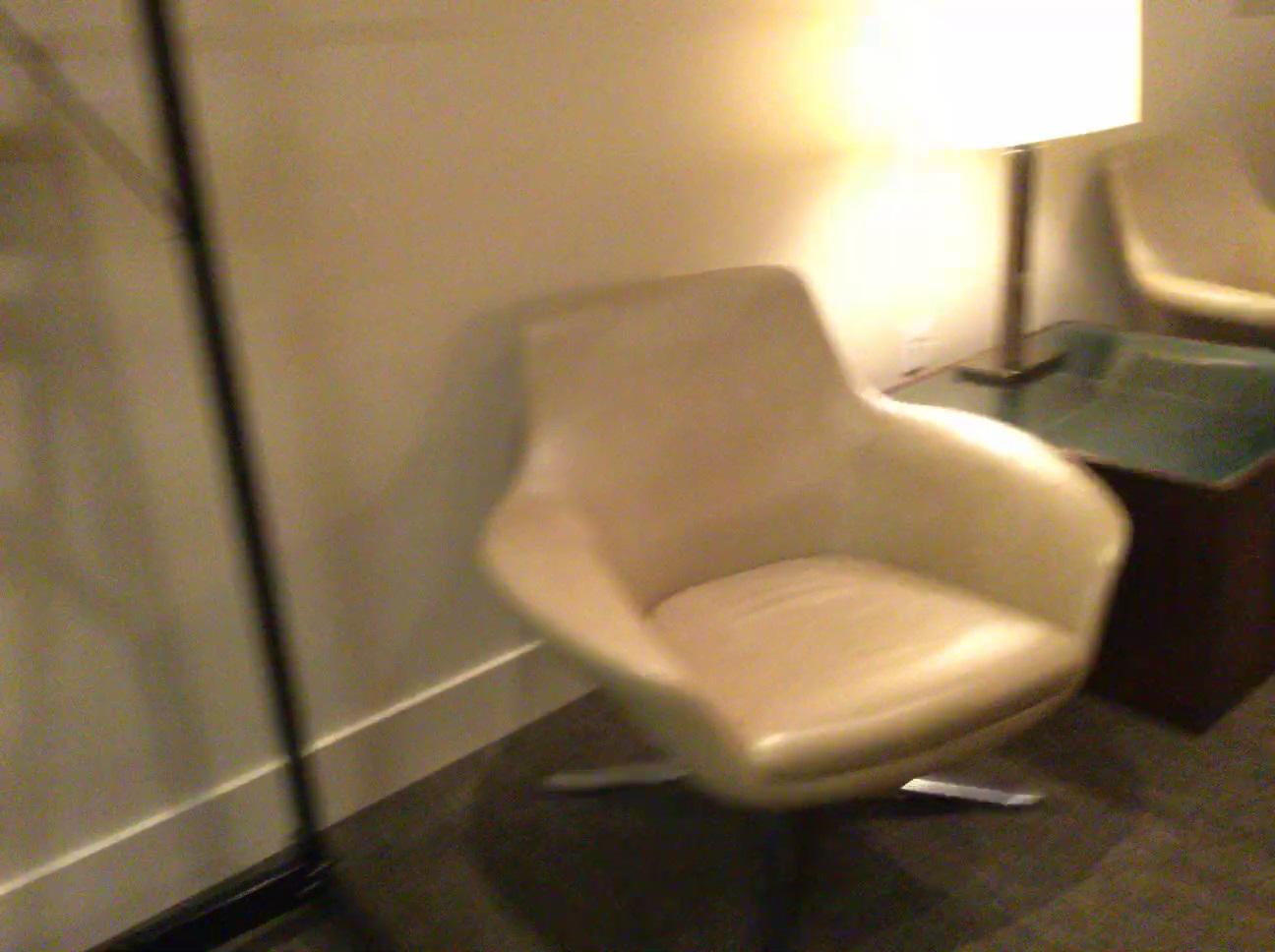}};
		\node[label={\textit{\small CNN}}] (net) at (-6.4, 0.0) {\includegraphics[width=0.15\linewidth]{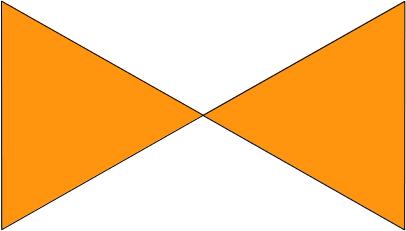}};
		\node[label={[label distance=-2mm]\textit{\small $O_1$}}] (segpred) at (-4.8, 0.0) {\includegraphics[width=0.2\linewidth]{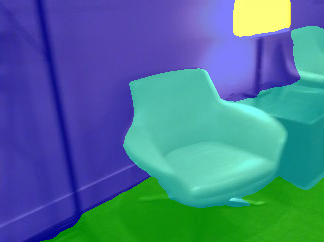}};
		
		\node[label={[label distance=-2mm]\textit{\small Annotation}}] (annot) at (-1.6, 0.0) {\includegraphics[width=0.2\linewidth]{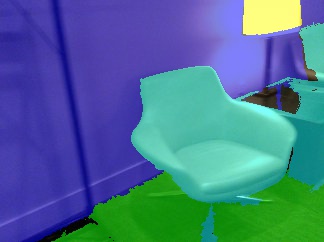}};

		\draw[-, line width=1mm, color=black] (simg) to[] (net);
		\draw[-, line width=1mm, color=black] (net) to[] (segpred);
		
		\node[label={}, circle,draw=black, line width=1pt, inner sep=0pt,minimum size=28pt] (ce) at (-3.2,0.0){$L_S$};
		
		\draw[dotted, line width=1mm, color=black] (segpred) to[] (ce);
		\draw[dotted, line width=1mm, color=black] (ce) to[] (annot);
		
		\node[label={[label distance=-2mm]\textit{\small $I_2$}}] (img1) at (-8.0, -1.7) {\includegraphics[width=0.2\linewidth]{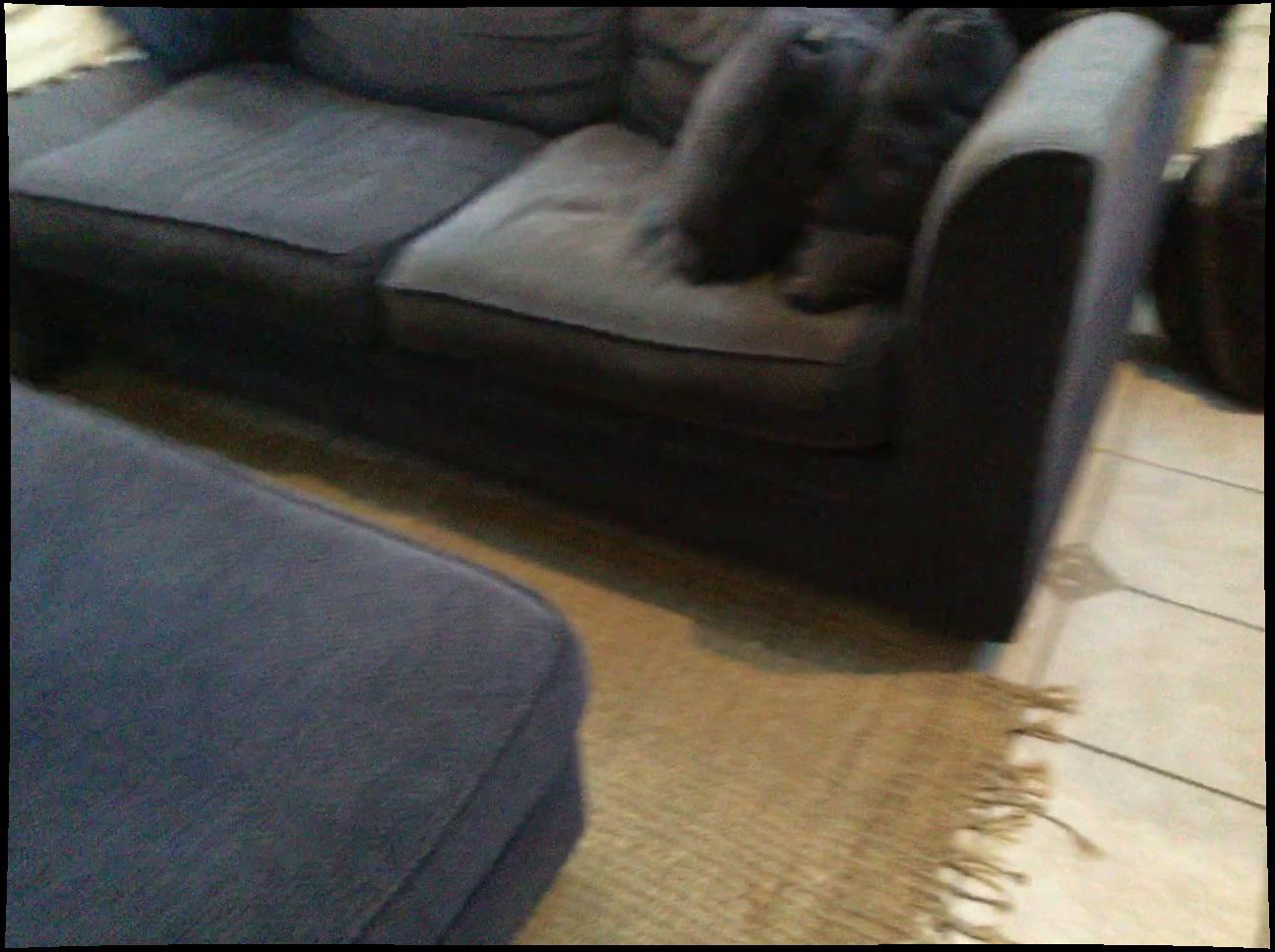}};
		
		\node[label={\textit{}}] (net1) at (-6.4, -1.7) {\includegraphics[width=0.15\linewidth]{figures/introduction/network/network}};
		
		\node[label={[label distance=-2mm]\textit{\small $O_2$}}] (out1) at (-4.8, -1.7) {\includegraphics[width=0.2\linewidth]{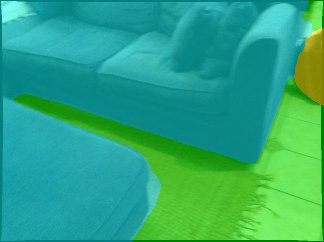}};
		
		\node[label={[label distance=-2mm]\textit{\small $I_3$}}] (img2) at (-8.0, -3.4) {\includegraphics[width=0.2\linewidth]{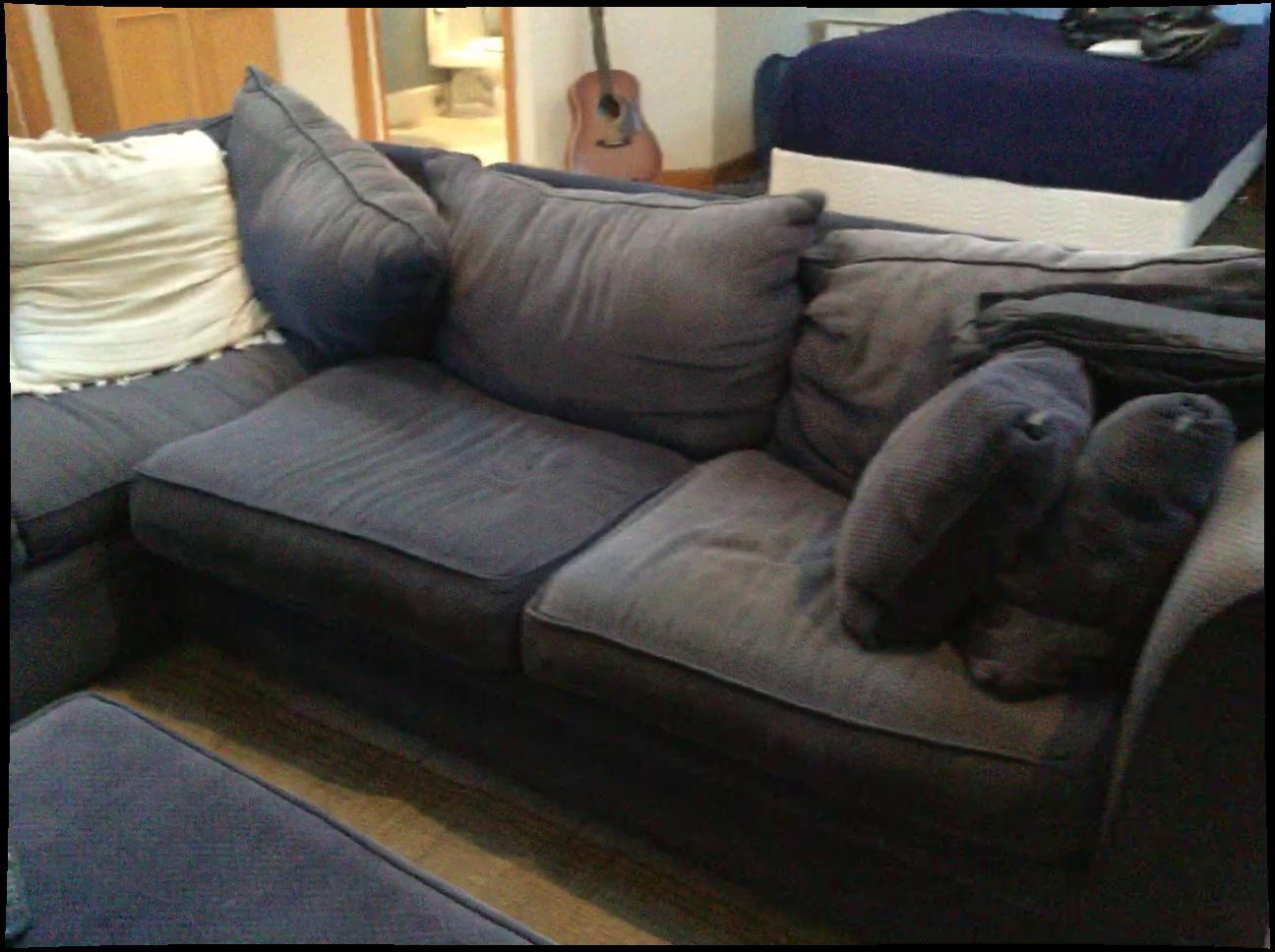}};
		
		\node[label={\textit{}}] (net2) at (-6.4, -3.4) {\includegraphics[width=0.15\linewidth]{figures/introduction/network/network}};
		
		\node[label={[label distance=-2mm]\textit{\small $O_3$}}] (out2) at (-4.8, -3.4) {\includegraphics[width=0.2\linewidth]{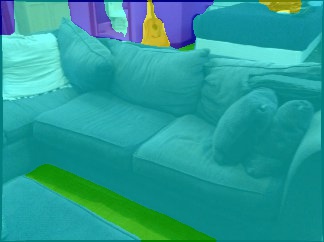}};
		
		\node[label=below:{\textit{\small Warped Output}}] (wout2) at (-1.6, -3.4) {\includegraphics[width=0.2\linewidth]{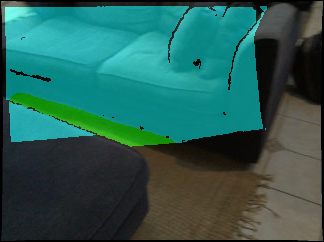}};
		
		\draw[-, line width=1mm, color=black] (img1) to[] (net1);
		\draw[-, line width=1mm, color=black] (net1) to[] (out1);
		
		\draw[-, line width=1mm, color=black] (img2) to[] (net2);
		\draw[-, line width=1mm, color=black] (net2) to[] (out2);
		
		\draw[dashed,->,line width=1mm, color=black] (out2)  to[bend right=30] node[pos=0.5, below]{} (wout2);
		
		\node[label=below:{}, circle,draw=black, line width=1pt, inner sep=0pt,minimum size=28pt] (l1) at (-3.2,-1.5){$L_G$};
		
		\draw[dotted,line width=1mm, color=black] (wout2)  to[] (-1.6, -1.5) to[] (l1);
		
		\draw[dotted,line width=1mm, color=black] (l1)  to[]  (out1);
		
		\draw [dotted, draw=red, line width=0.5mm] (-3.9,-2.6) rectangle
		(-2.5,1.0) node[] at (-2.9,1.2){};
		
		\node at (-3.2,-0.75){\Large +};
		
		\draw[<->, line width=0.5mm, color=black] (net) to[] (net1);
		
		\draw[<->, line width=0.5mm, color=black] (net1) to[] (net2);
		
		\draw[dashed, line width=0.5mm, color=black] (-6.4,-3.6) -- node[label={[label distance=1mm]below: \textit{\small Shared weights}}]{} (-6.4,-4.2);
		
		

		\end{tikzpicture}
		\caption{Training: We exploit some manual labels, as for Image
			$I_1$, in a standard supervised fashion, and geometric
			constraints, as between Images $I_2$ and $I_3$: The labels
			predicted for Image $I_2$ should be consistent with the labels
			predicted for Image $I_3$.}
	\end{subfigure}
	
	\vspace{0.5cm}
	
	\begin{subfigure}[b]{0.9\linewidth}
		\hspace{-0.6cm}
		\begin{tabular}{cccc}

			\rotatebox{90}{$\>$Supervised} &
			\includegraphics[width=0.3\linewidth]{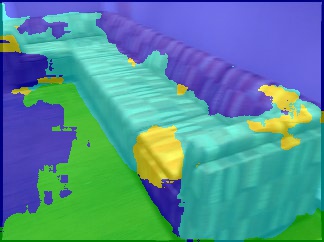}
			&
			\includegraphics[width=0.3\linewidth]{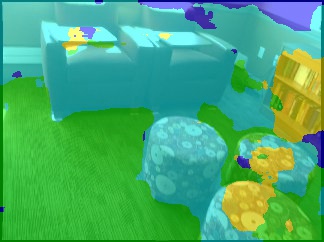}
			&
			\includegraphics[trim={0.3cm 0.7cm 2.7cm 1.5cm},clip,width=0.3\linewidth]{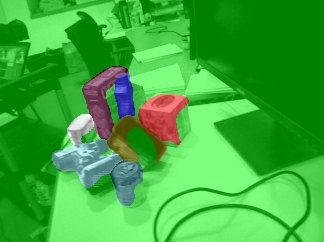}
			\\
			\rotatebox{90}{$\quad$S4-Net} &
			\includegraphics[width=0.3\linewidth]{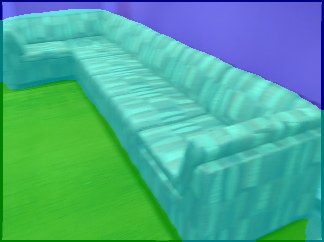}
			&
			\includegraphics[width=0.3\linewidth]{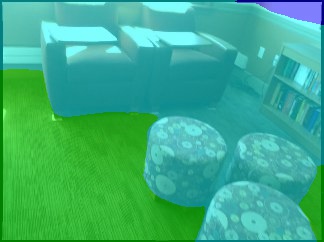}
			&
			\includegraphics[trim={0.3cm 0.7cm 2.7cm 1.5cm},clip,width=0.3\linewidth]{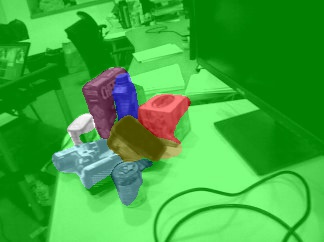}\\
		\end{tabular}
		
		\caption{Testing:  Segmentation  from   a  single  color  image.
			Because  we can  exploit many  training sequences  without any
			labels, we can outperform supervised learning.}
	\end{subfigure}
	\vspace{0.2cm}  
	\caption{Our approach exploits some manual labels and geometric consistency on
		image sequences  to train  a network to  perform semantic  segmentation.  At
		test time,  the network predicts  a dense  segmentation from a  single color
		image.}
	\label{fig:abstract_approach}
\end{figure}

Semantic segmentation  of images provides  high-level understanding of  a scene,
which is useful  in many applications including robotics  and augmented reality.
Recent approaches  rely on deep learning  and large amount of  data specifically
annotated for  the task at  hand~\cite{Long2015, Dai2017, HeGDG17}, but,  as for
many other  computer vision problems,  getting such annotations is  a cumbersome
process.  A  popular solution is crowdsourcing  that can spread the  efforts for
annotating  data   significantly~\cite{Dai2017}.   Nonetheless,   it  introduces
additional    expenses  and   the  resulting  annotations   produced  by
non-experts may not be satisfying.

An alternative approach  is to reduce the amount of  required manual annotations
for learning.  One way  of doing this is through data  augmentation which can be
achieved           by           performing           appropriate           image
transformations~\cite{RonnebergerFB15}. A  different option is to  use synthetic
data~\cite{RichterVRK16,  AlhaijaMMGR18}.   However,   creating  synthetic  data
requires the cumbersome  task of building realistic 3D  models.  Such approaches
hence introduce additional difficulties as  the domain gap between synthetic and
real data needs to be bridged.

In this  paper, we show  how semantic segmentation learning  can be cast  into a
semi-supervised framework.  A standard  technique from 'general' semi-supervised
learning is to  add constraints on pairs of unlabeled  training samples that are
close to each other,  enforcing the fact that such two  samples should belong to
the  same category.   In  the case  of semantic  segmentation  learning, we  can
introduce  similar semi-supervised  constraints, enforcing  the fact  that image
locations that correspond to the same 3D point should have the same label. As we
show  in Figure~\ref{fig:abstract_approach},  together with  standard supervised
terms from the  labeled images, these constraints result in  a powerful learning
process exploiting  the 3D geometry  nature of  the problem.  In  fact, starting
from RGB-D sequences with  a few annotated frames, our results  show that we can
keep  incorporating RGB-D  sequences without  any manual  labels to  improve the
overall  performance.   This  makes  our approach  very  convenient  since  such
sequences only  need to be captured,  without the effort of  manually annotating
them.

To demonstrate this idea, we make use  of RGB-D image sequences of rigid scenes,
and a  very small number  of manually annotated  frames.  We use  such sequences
because  they  provide  the  geometric information  required  to  introduce  our
semi-supervised constraints, while being  easily acquired with suitable cameras,
and automatically registered using  robust geometric algorithms.  Alternatively,
automated dense SLAM systems could be used to acquire such information for rigid
scenes   with  a   color   camera~\cite{Godard17,ZhouBSL17,Bloesch18},  or   for
deformables scenes with an RGB-D camera~\cite{Newcombe15, WhelanSGDL16}.

In short, our  goal is not to  develop a new semantic segmentation  method or to
improve performance on existing benchmarks, but to show that when multiple views
of  the  same scene  are  available, enforcing geometric  constraints  with
semi-supervised  learning   is  a  powerful  framework   for  learning  semantic
segmentation when only a small fraction of the available data is annotated.

We demonstrate  our semi-supervised training pipeline  on different segmentation
problems.       We      considered      the      ScanNet~\cite{Dai2017}      and
LabelFusion~\cite{Marion2018}  datasets  because  they provide  the  input  data
suitable to our approach.  First, we  apply our method to a 4-class segmentation
problem derived from  the ScanNet dataset.  For the experiments,  we compare our
method to a supervised approach. While  supervised approach is trained using all
of the available  annotations, we show that our approach  can achieve comparable
results even after  reducing the amount of manual annotations  for training to a
small  fraction.   Second, we  apply  our  method  to  segmenting objects  in  a
cluttered  scene  where many  partial  occlusions  are present  and  annotations
typically include only parts of the objects.  We show that with our method, only
one manually labeled image is sufficient for segmenting a complete scene.

\section{Related Work}

In this  section, we  briefly discuss  related work on  the aspects  of semantic
segmentation,  general semi-supervised  learning,  and also  recent methods  for
learning depth  prediction from  single images, as  they also  exploit geometric
constraints similar to our approach.

\subsection{Supervised Semantic Segmentation with Deep Networks}

Supervised semantic segmentation algorithms rely  on principles that for a given
image, the algorithm learns to output segmentation prediction that is similar to
its ground truth annotation. In contrast to classification tasks, where a single
label is required per image, semantic  labels are on pixel-wise level.  Prior to
deep learning,  Conditional Random  Fields~\cite{ReynoldsM07, PlathTN09}  were a
popular method for tackling this task.

Introduction of deep  learning has made a  large impact on the  task of semantic
segmentation.    Fully   Convolutional  Networks~(FCN)~\cite{Long2015}   allowed
segmentation prediction for  input of arbitrary size. In  this setting, standard
classification task networks~\cite{SimonyanZ15a, HeZRS16} are utilized and fully
connected layers transformed into convolutions.  FCNs use deconvolutional layers
that  learn the  interpolation  for upsampling  process.  Other works  including
SegNet~\cite{BadrinarayananK17},        U-Net~\cite{RonnebergerFB15},        and
DeepLab~\cite{ChenPKMY18} rely  on similar architectures.  Such  works have been
applied   to   a    variety   of   segmentational   tasks~\cite{RonnebergerFB15,
  ArmaganHRL17, MiliotoLS18}.

In our experiments, we use the U-Net architecture, because it is simple yet very
powerful.  However, any other architecture could be used instead.

\subsection{Semi-Supervised Learning with Deep Networks}

The main  limitation of supervised methods  is the availability of  ground truth
labels.  In contrast,  semi-supervised learning is a general  approach aiming at
exploiting both  labeled and unlabeled  training data.  Some approaches  rely on
adversarial    networks    to    measure   the    performance    of    unlabeled
data~\cite{DaiYYCS17, SantosWZ17, Kumar17, Hung2018}. More in line with our work
are   the   popular   consistency-based   models~\cite{LaineA17,   TarvainenV17,
  athiwaratkun2018}.  These  methods enforce the  model output to  be consistent
under  small  input  perturbations.   As  explained  in~\cite{athiwaratkun2018},
consistency-based models  can be viewed  as a student-teacher model:  To measure
the consistency of a model $f$, or  the student, its predictions are compared to
the predictions of a teacher model $g$,  a different trained model, while at the
same time applying small input perturbations.

The   $\Pi\mathit{-model}$~\cite{LaineA17}   is   a  recent   method   using   a
consistency-based  model  where the  student  is  its own  teacher,  \emph{i.e.}
$f=g$.  It relies on a cross-entropy  loss term applied to labeled data
only and an additional term that  penalizes differences in predictions for small
perturbations of input data.  Our semi-supervised approach is closely related to
the $\Pi\mathit{-model}$  but relies on  geometric consistency instead  of the
input perturbations.

\subsection{Single-View Depth Estimation}

Because  of  view warping,  our  approach  is also  related  to  recent work  on
unsupervised single-view depth estimation.  Both Zhou~\etal~\cite{ZhouBSL17} and
Godard~\etal~\cite{Godard17}  proposed  an  unsupervised approach  for  learning
depth estimation from  video data. This is  done by learning to  predict a depth
map so  that a  view can be  warped into another  one.  This  research direction
became   quickly    popular,   and   has    been   extended   since    by   many
authors~\cite{Yin18, Mahjourian18, Wang18, Godard18}.

Our work is  related to these methods as it  also introduces constraints between
multiple views, by  using warping. However, since we focus  on semantics and not
geometry, an  input from the  user is still  required to indicate  the different
categories.

\section{Method}

\newcommand{\calS}{\mathcal{S}}
\newcommand{\calU}{\mathcal{U}}
\newcommand{\calW}{\mathcal{W}}
\newcommand{\calWP}{\mathcal{\WP}}
\newcommand{\WA}{W\!\!A}
\newcommand{\WP}{W\!\!P}
\newcommand{\NV}{N\!\!V}
\newcommand{\Warp}[2]{\text{Warp}_{#1\rightarrow#2}}
\newcommand{\calN}{\mathcal{N}}
\newcommand{\CE}{{\text{CE}}}

We assume we are given a small set  of registered images and their corresponding
depths captured in one or several  scenes and annotated by a user:
\[
\calS = \{e_i = (I_i, A_i, D_i, T_i, q_i)\}_i \> ,
\]
where $I_i$  is a  color image, $A_i$  is its manual  annotations, $D_i$  is its
depth map, $T_i$ is the corresponding camera pose, and $q_i$ is the index of the
sequence.  We are also  given a set of similar examples, but  for which there is
no available manual annotations:
\[
\calU = \{e_j = (I_j, D_j, T_j, q_j)\}_j \> .
\]
The samples in $\calU$ can be from  the same sequences as the samples in $\calS$
but we  can also have samples  in $\calU$ from  sequences that do not  appear in
$\calS$. In other  words, we  can have  scenes for  which no  manual labels  are
available and only geometric consistency can be exploited.

We would like  to train a network  $f()$ using these data to  segment views from
novel scenes using only a color image as input.

\subsection{Automatic Warping of the Manual Annotations}
\label{sec:aug}

As shown  in Fig.~\ref{fig:seg_synth},  by warping  the manual  annotations from
$\calS$ to the images  in $\calU$ from the same sequences,  we can obtain easily
additional  annotations  for these  images.  In  other  terms, we  can  generate
annotations $A_i$ for some of the images  in $\calU$, provided they are from the
same sequence as a sample in $\calS$ and overlap its point of view.

\emph{We  consider these  additional annotations  like the  manual annotations},
even if they  are obtained automatically, and  we will use them  in a supervised
way.  However,  these annotations are only  partial: Parts of images  in $\calU$
will remain unlabeled if they are not seen in images in $\calS$, and some images
in $\calU$ will remain completely unlabeled if  they do not belong to a sequence
present  in  $\calS$.  These  unlabeled  parts  can  then  be exploited  by  our
semi-supervised terms.

\begin{figure}
\centering
\begin{tikzpicture}
\node[label={\textit{\small Manual annotation}}] (neighb) at (0,0) {	\includegraphics[width=0.4\linewidth]{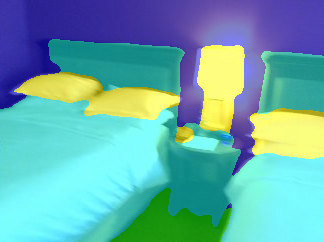}};

\node[label={\textit{\small Manual annotation after warping}}] (merged) at (4,0) {	\includegraphics[width=0.4\linewidth]{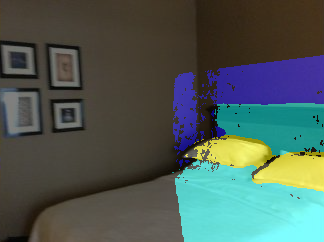}};
\draw[->, line width=4pt, color=orange] (neighb) to[bend right=45] (merged);
\end{tikzpicture}
\caption{By warping  the manual  annotations to the  non-labeled images,  we can
  easily   obtain  additional   annotations.   We   consider  these   additional
  annotations like  the manual annotations and  use them in the  supervised loss
  term. Nevertheless, the semi-supervised geometric consistency term
  still has a significant benefit. }
\label{fig:seg_synth}
\end{figure}

In practice,  due to errors  in depth  maps, registration, or  manual annotation
mistakes, the manual  annotations $A_i$ may be in conflict  with each other when
warped to  the same  image: For  a pixel  of an  image in  $\calU$, we  may have
several possible classes.  If  it is the case, we keep  the most frequent class.
If there  are still several  possible classes,  we select one  randomly, however
this case is extremely rare in practice.



\subsection{Semi-Supervised Learning for Semantic Segmentation}
\label{sec:sslss}

To  train $f$,  we  propose to  minimize  the following  loss  function for  our
semi-supervised approach over its parameters $\Theta$:
\begin{equation}
  L = L_S + \lambda L_G \> ,
  \label{eq:loss}
\end{equation}
where  $L_S$ is  a loss  term  for supervised  learning,  $L_G$ is  a loss  term
introducing  our   geometric  consistency   constraints,  and  $\lambda$   is  a
regularization  factor.   For $L_S$,  we  use  a  standard term  for  supervised
semantic segmentation:
\begin{equation}
L_S = \sum_{e\in\calS\cup\calU} l_\CE(f(I(e);\Theta), A(e)) \> ,
\end{equation}
with $l_{\text{CE}}$ the cross-entropy function comparing the network prediction
$f(I;\Theta)$ for  image $I$ and the  manual annotation $A$.  $I(e)$  and $A(e)$
simply  denote  the  image  and  annotation for  sample  $e$.  As  explained  in
Section~\ref{sec:aug},  for  the  samples  in $\calU$  we  use  the  annotations
obtained automatically. There are samples in  $\calU$ with no annotations and we
simply ignore them.


To exploit  unlabeled images and  geometric consistency, we introduce  the $L_G$
loss term, which penalizes the differences in predicted probabilities:
\begin{equation}
L_G = \sum_{e\in\calU} \sum_{e' \in \calN(e)} | f(I(e); \Theta) - \Warp{e'}{e}(f(I(e'); \Theta)) | \> ,
\label{eq:L_G}
\end{equation}
where $\calN(e)$ is a set of samples with a point of view that overlaps with the
point of  view of  $e$.  In  practice, we simply  use the  frames from  the same
sequence as $e$.  $\Warp{e'}{e}(P')$ warps labels $P'$ for sample $e'$ to the view
for sample $e$.  More details about  the $\text{Warp}$ function are given below.
We consider prediction $f(I(e'); \Theta)$ as a teacher prediction and, similarly to the $\Pi\mathit{-model}$~\cite{LaineA17}, it is treated as a constant when calculating the update of the network parameters. We use the $\ell_1$  norm to compare the predicted probabilities.




\comment{
  
\begin{figure}
\centering
\begin{tikzpicture}
\node[label={\textit{\small Predicted Segmentation}}] (neighb) at (0,0) {	\includegraphics[width=0.4\linewidth]{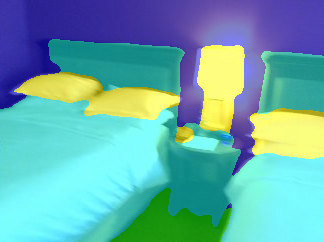}};

\node[label={\textit{\small Synthesized Segmentation}}] (merged) at (4,0) {	\includegraphics[width=0.4\linewidth]{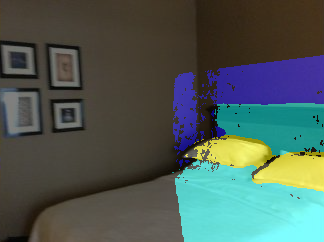}};
\draw[->, line width=4pt, color=orange] (neighb) to[bend right=45] (merged);
\end{tikzpicture}
\caption{Synthesized segmentation provides learning signal for unannotated frames.}
\label{fig:seg_synth}
\end{figure}

}


\subsection{Label Warping}


To implement the warp function $P = \Warp{e'}{e}(P')$, we rely on the method used
in \cite{ZhouBSL17}  for intensity image  warping. $P'$  is a set  of probability
maps, one for each possible label, for  a source sample $e'$. The warp function
provides a probability map $P$ for the labels of the target sample $e$, obtained by warping $P'$ using
the depth map  for $e'$ and the rigid  motion between $e'$ and $e$.   Given the 2D
location $p$ in homogeneous coordinates and the  depth $d$ of a pixel for the
target sample, we  can compute the 2D location $p'$  of the corresponding 2D
point in the source sample $e'$:
\begin{align}
p' = K T_{e \rightarrow e'} d K^{-1} p \> ,
\end{align}
where $K$ is the matrix of  camera internal parameters, $T_{e \rightarrow e'}$ is
the rigid motion from  the target sample to the source  sample.  Since $p$ usually
lies  between  integer  pixel  locations, we  use  the  differentiable  bilinear
interpolation from \cite{JaderbergSZK15} to  compute the final probabilities for
$P'(p')$ from the 4 neighbouring pixels.

The advantage of this transformation is that it is differentiable, and can thus
be used in the loss function of Eq.~\eqref{eq:loss} to train the network.

In practice,  not every  pixel in the  target sample has  a correspondent  in the
source sample, and we ignore them in  the loss function.  This can happen because
depth is not necessarily available for every pixel when using depth cameras, and
because some pixels  in the target sample  may not be visible in  the source sample,
because they  are occluded or, simply,  because they are  not in the field  of view of the
source sample. We detect if a pixel is  occluded by comparing the original
source depth at the mapped location to the transformed target depth given by the
third coordinate of $p'$.  If the  difference between the depths is large, this
means that the pixel is occluded and does not correspond to the same physical 3D
point.



\subsection{Network Initialization}
\label{sec:init}
To initialize  the network $f(.;\Theta)$,  we first train  it only based  on the
$L_S$ loss term. This  avoids the problem of converging to  a bad local minimum
introduced by  the term $L_G$.  As it is  the case with  other consistency-based
models,  minimizing  $L_G$ may  fall  in  a solution  where  a  single class  is
predicted  for all  the  image locations.   Even  though tuning  hyper-parameter
$\lambda$ more carefully might resolve this  problem, we noticed that using this
pre-training step makes the convergence to a correct model easier.


\subsection{Network Architecture}

We  implement the  network $f$  as a  U-Net architecture~\cite{RonnebergerFB15},
with $5$ layers for both the encoding  and decoding parts.  For the encoder, the
number  of  features in  the  first  layer is  $32$  and  doubles up  for  every
additional  layer,  and  the  decoder  is  symmetric  to  the  encoder.  We  use
convolutional  filters of  size $3$  and apply  reflection padding.  Max-pooling
layers use regions of  size $2$. All layers use ReLU  activations except for the
output  layer,  which  uses  the  Softmax  activation  to  predict  segmentation
probabilities.  Other  architectures could be used  but U-Net proved to  be both
convenient and powerful enough for our purpose.

\section{Evaluation}
\label{sec:eval}

We evaluate our approach we call  S4-Net on two different segmentation problems.
We use the ScanNet dataset~\cite{Dai2017}  and evaluate our approach on a
4-class segmentation problem demonstrating the influence of the number of annotated images
on the final performance.  Additionally, we evaluate the performance of S4-Net on the
task of segmenting objects on the LabelFusion dataset~\cite{Marion2018}.


\textbf{Training Details:} For  all experiments, we set the  $\lambda$ factor in
the loss  function to $0.1$.   For different loss  terms we use  different batch
sizes.  The  batch size  for $L_S$  is set  to $4$.  Since $L_G$  predicts larger
number of segmentations for each iteration,  we set the corresponding batch size
to $1$.  We  use the Adam optimizer~\cite{KingmaB15} with  initial learning rate
of $10^{-4}$  and train  the network  until convergence.   The input  images are
resized to $320 \times 242$.  As mentioned in Section~\ref{sec:init}, for better
convergence, we pre-train the network only using the $L_S$ term.


\subsection{Segmenting 4 Classes on ScanNet}

For our first experiment, we evaluate our method on \textit{structural classes},
as defined in the NYU-Depth dataset~\cite{SilbermanHKF12}. Objects are
classified to reflect their physical role in the scene: \textit{ground},
\textit{permanent structures}~(structures that do not move such as walls and
ceilings), \textit{furniture} and \textit{props}~(easily movable objects).

Unfortunately,  the NYU-Depth  dataset is  small and  the camera  poses are  not
provided.   We therefore  turned to  the ScanNet  dataset~\cite{Dai2017}. It  is
organized    into    different    scenes     and    for    each    scene,    and
BundleFusion~\cite{Dai2017bundlefusion} was  used to register the  RGB-D images.
Furthermore,  the dataset  provides annotated  3D reconstructions  of individual
scenes  and respectively  the annotation  mappings  for each  of the  individual
frames.  Since the  ScanNet dataset  provides annotation  mappings to  NYU-Depth
segmentation  classes,  we  did  not  have  to  create  any  manual  annotations
ourselves.

As this dataset was annotated by a crowd sourcing community, the same types of objects can be annotated with different classes across the different scenes, or might not be annotated at all in some cases. Mapping to a 4-class semantic segmentation problem helps to alleviate this issue, but does not solve it completely.


\renewcommand{\arraystretch}{0.9} 
\newcommand{\sm}[1]{{\small #1}}

\begin{figure*}
\begin{tabular}{c}

\begin{tabular}{@{}lcccccccrcccccc@{}}
  \toprule
  & \phantom{}&  \multicolumn{6}{c}{{\bf Accuracy}} & \phantom{a}& \multicolumn{6}{c}{{\bf Intersection over Union}}\\
  \cmidrule{3-8} \cmidrule{10-15}
  && \multicolumn{6}{c}{{\small annotation percentage (\%)}} && \multicolumn{6}{c}{{\small annotation percentage (\%)}} \\
  && \sm{0.1} & \sm{0.25} & \sm{0.5} & \sm{1} & \sm{2} & \sm{100} &&  \sm{0.1} & \sm{0.25} & \sm{0.5} & \sm{1} & \sm{2} & \sm{100}   \\
  \midrule
  SL w/o annot. warping    && 64.2 & 68.3 & 73.6 & 81.3 & 88.3 & 98.6  && 37 & 40.2&43.4&51.4&60.4&92.9 \\
  SL w/ annot. warping     && 82.4 & 87.1 & 94.2 & 96.7 & 97.2 & n/a  && 58 & 65.2 & 78.6 & 85.9 & 87.3 & n/a\\
  S4-Net                   && 86.7 & 90.4 & 95.3 & 97 & 97.2 & n/a && 68.6&76.7&82.6&86.9&89&n/a \\
  \bottomrule
\end{tabular}
\\
(a) Accuracy and Intersection-over-Union metrics on the training set.\\
\\
\begin{tabular}{@{}lcccccccrcccccc@{}}
  \toprule
  & \phantom{}&  \multicolumn{6}{c}{{\bf Accuracy}} & \phantom{a}& \multicolumn{6}{c}{{\bf Intersection over Union}}\\
  \cmidrule{3-8} \cmidrule{10-15}
  && \multicolumn{6}{c}{{\small annotation percentage (\%)}} && \multicolumn{6}{c}{{\small annotation percentage (\%)}} \\
  && \sm{0.1} & \sm{0.25} & \sm{0.5} & \sm{1} & \sm{2} & \sm{100} &&  \sm{0.1} & \sm{0.25} & \sm{0.5} & \sm{1} & \sm{2} & \sm{100}   \\
  \midrule
  SL w/o annot. warping    && 60 & 63.1 & 66.3 & 70.6 & 75.7 & 84.9  && 33.2 & 35.7 & 37.1 & 41.6 & 46.1 & 60.8  \\
  SL w/ annot. warping     && 76 & 78.6 & 83.5 & 84.4 & 84.8 & n/a  && 48.3 & 51.1 & 57.9 & 59.8 & 59.5 & n/a \\
  S4-Net                   && 78.9   & 82.8 & 84.5 & 84.7 & 82 & n/a && 52.3 & 56.9 & 60.4 & 59.9 & 62.7 & n/a \\
  \bottomrule
\end{tabular}
\\
(b) Accuracy and Intersection-over-Union metrics on the test set.\\
\\
\begin{tabular}{@{}lcccccccrcccccc@{}}
  \toprule
  & \phantom{}&  \multicolumn{6}{c}{{\bf Accuracy}} & \phantom{a}& \multicolumn{6}{c}{{\bf Intersection over Union}}\\
  \cmidrule{3-8} \cmidrule{10-15}
  && \multicolumn{6}{c}{{\small annotation percentage (\%)}} && \multicolumn{6}{c}{{\small annotation percentage (\%)}} \\
  && \sm{0.1} & \sm{0.25} & \sm{0.5} & \sm{1} & \sm{2} & \sm{100} &&  \sm{0.1} & \sm{0.25} & \sm{0.5} & \sm{1} & \sm{2} & \sm{100}   \\
  \midrule
  SL w/o annot. warping    && 56.1 & 56.5 & 60.8 & 61   & 63.2 &  66.2 &&  27 & 27.1 & 29.1 & 30.4 & 30.9 & 33.4\\
  SL w/ annot. warping     && 60.7 & 63.7 & 63.9 & 66.8 & 64.3 &  n/a  && 30.3 & 32.5 & 31.6 & 35 & 33.4 & n/a\\
  S4-Net                   && 63   & 67.6 & 65.4 & 66.7 & 66   &  n/a  && 32.2 & 35.6 & 34.5 & 34.5 & 34.1 &n/a\\
  S4-Net w/ n-a sequences  && 65.3 & 73.1 & 69.7 & 72.4 & 73.2 &  n/a  &&  37 & 49.5 & 43.3 & 46.4 & 49.8 &n/a\\
  \bottomrule
\end{tabular}
\\
(c) Accuracy and Intersection-over-Union metrics on the 'generalization test set'.\\
\\
\begin{tabular}{cccc}
\includegraphics[width=1.\linewidth]{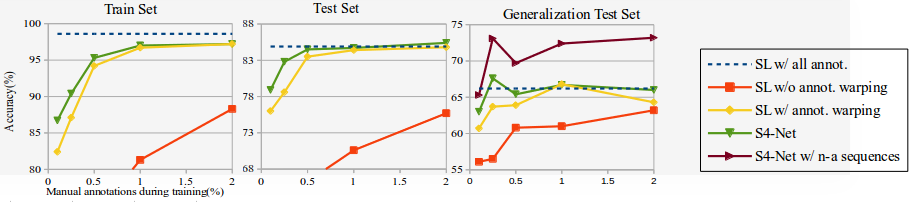}&\\

\end{tabular}
\\
(d) Accuracy metric on different datasets for the different methods
when varying the percentage of annotated images.\\
\begin{tabular}{cccc}
\includegraphics[width=1.\linewidth]{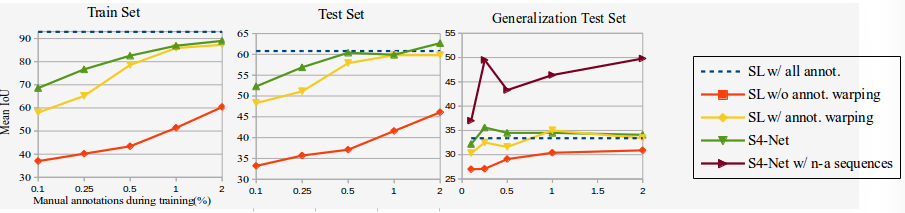}&\\
\end{tabular}
\\
(e) Same for the Intersection-over-Union metric.\\
\end{tabular}

\caption{\label{fig:eval_graph}Datasets created  from ScanNet, S4-Net average  performances in respect
  to  the  percentage of  annotations  used  during  training. 'SL'  stands  for
  supervised learning,  and 'S4-Net  w/ n-a sequences'  for S4-Net  trained with
  additional non-annotated  sequences.  The  performances on  the 'generalisation
  test set' are the ones relevant for real applications.}
\end{figure*}

In this experiment, we  evaluate the behavior of S4-Net as we  vary the number of
manual annotations that  are used for training.  For each  scene in the training
set, we therefore discard uniformly desired  number of annotations and apply our
approach. We compare S4-Net to the  supervised learning approach with and without
annotation warping in Figure~\ref{fig:eval_graph}.  More precisely, we use the following sets:
\begin{itemize}
	\item The training set consists of the first 26 scenes from ScanNet.
	\item The test set consists of alternative scans of the scenes seen in the training set.
	\item The 'generalization test set' consists of scenes that were not in the training set. The performances on this set are the performances relevant for practical applications.
\end{itemize}

Additionally, during training, we use  a validation set that includes alternative
scans of the training scenes that are not in the test set nor in the generalization
test  set.  It is  used  for  validation  of  network parameters  for  different
iterations during  training.  At the end  of training, we keep  the network with
the  highest validation  accuracy for  further evaluation.

The  performances are  measured  with  the accuracy  of  the network  prediction
compared to the provided ground truth annotations:
\[ 
Acc = \frac{1}{\#images} \sum_{image} \frac{\#CorrectPredictions}{\#ValidAnnotations} \> ,
\]
and the Intersection over Union~(IOU) score.

Supervised learning with warped annotations significantly outperforms supervised
learning  without these  annotations.  Still,  our  S4-Net approach  is able  to
improve the performances even further.

\begin{figure*}
	\centering
	\begin{subfigure}[b]{.3\linewidth}
		\includegraphics[width=1.\linewidth]{figures/evaluation/testgen_unsupervised/scene0027_00/sup_seg_00548}
	\end{subfigure}
	\begin{subfigure}[b]{.3\linewidth}
		\includegraphics[width=1.\linewidth]{figures/evaluation/testgen_unsupervised/scene0027_00/ss_seg_00548}
	\end{subfigure}
	\begin{subfigure}[b]{.3\linewidth}
		\includegraphics[width=1.\linewidth]{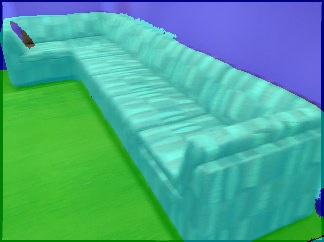}
	\end{subfigure}
	
	\begin{subfigure}[b]{.3\linewidth}
		\includegraphics[width=1.\linewidth]{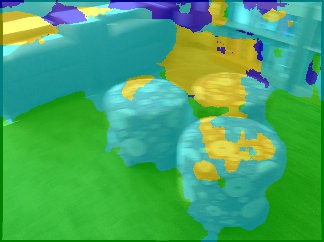}
	\end{subfigure}
	\begin{subfigure}[b]{.3\linewidth}
		\includegraphics[width=1.\linewidth]{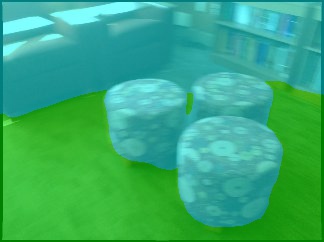}
	\end{subfigure}
	\begin{subfigure}[b]{.3\linewidth}
		\includegraphics[width=1.\linewidth]{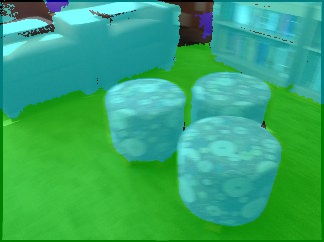}
	\end{subfigure}
	
	\begin{subfigure}[b]{.3\linewidth}
		\includegraphics[width=1.\linewidth]{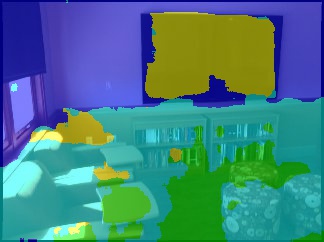}
	\end{subfigure}
	\begin{subfigure}[b]{.3\linewidth}
		\includegraphics[width=1.\linewidth]{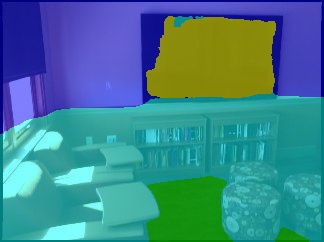}
	\end{subfigure}
	\begin{subfigure}[b]{.3\linewidth}
		\includegraphics[width=1.\linewidth]{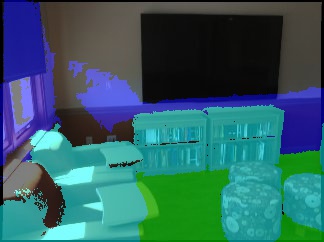}
	\end{subfigure}
	
	\begin{subfigure}[b]{.3\linewidth}
		\includegraphics[width=1.\linewidth]{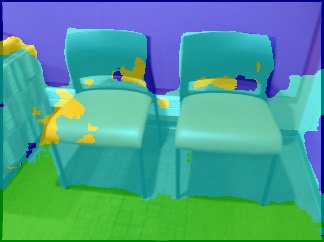}
	\end{subfigure}
	\begin{subfigure}[b]{.3\linewidth}
		\includegraphics[width=1.\linewidth]{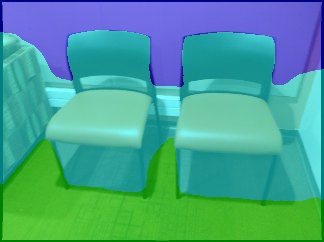}
	\end{subfigure}
	\begin{subfigure}[b]{.3\linewidth}
		\includegraphics[width=1.\linewidth]{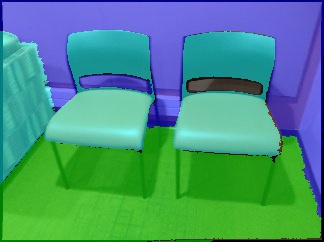}
	\end{subfigure}
	
	\begin{subfigure}[b]{.3\linewidth}
		\includegraphics[width=1.\linewidth]{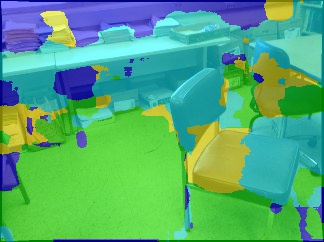}
		\caption{Supervised approach}
	\end{subfigure}
	\begin{subfigure}[b]{.3\linewidth}
		\includegraphics[width=1.\linewidth]{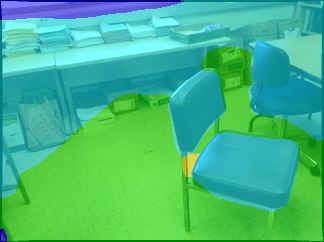}
		\caption{S4-Net}
	\end{subfigure}
	\begin{subfigure}[b]{.3\linewidth}
		\includegraphics[width=1.\linewidth]{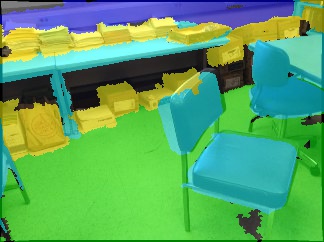}
		\caption{Ground Truth}
	\end{subfigure}
	\caption{\label{fig:gen_test_examples}\label{fig:gen_examples}Qualitative results on the generalization test set created from
          ScanNet.  The supervised  approach is able to  predict correctly large
          parts  of  the  segmentations  on  new scenes  but  still  makes  many
          mistakes.   S4-Net can  exploit  additional  non-labeled sequences  of
          other scenes  to correct most of  these mistakes.  The floor  class is
          shown  in   {\color{floor}\textbf{green}},  the  structure   class  in
          {\color{structure}    \textbf{blue}},   the    furniture   class    in
          {\color{furniture}\textbf{cyan}},    and    the   props    class    in
          {\color{props}\textbf{yellow}}. }
\end{figure*}

Finally, we show  the ability of S4-Net to learn  from additional scenes without
any annotations.  The  annotated set $\calS$ consists of  all manual annotations
from the  first 26 scenes.   For the non-annotated set,  we use scenes  from the
generalized test set  \emph{i.e.} scenes that appear neither in  $\calS$ nor in
the test set.  As shown in Table~(c) of Figure~\ref{fig:eval_graph}, this
significantly  improves  the performance.   Figure~\ref{fig:gen_examples}  shows
some  qualitative results  to  give  a visual  idea  of  this improvement.   The
supervised approach with annotation warping  often makes many small mistakes and
enforcing geometric constraints  during training of S4-Net managed  to fix these
mistakes.

\subsection{Object Segmentation}

For this experiment, we  consider the LabelFusion dataset~\cite{Marion2018}.  It
contains  RGB-D recordings  of  multi-object scenes  and corresponding  accurate
poses,  obtained  through  the  ElasticFusion  SLAM  system~\cite{WhelanSGDL16}.
LabelFusion  provides  accurate  segmentations  of  the  objects,  which  allows
evaluating the segmentation results.

Given a  sequence, we chose a  single image that  captures most of a  scene, and
used the annotation for this image for training.  For this problem, we observed
that standard  cross-entropy for the  supervised loss function  performed poorly
because most of the scene points belong to the background class, and we replaced
it  by  the  class-weighted cross-entropy~\cite{PanchapagesanSK16}.   This  loss
function down-weights  the errors in  prediction for frequently  appearing classes
and up-weights  the errors for  infrequently appearing classes.  Hence,  we only
down-weight the  errors for  the background  class to $0.1$.   This is  the only
change we make in respect to original definitions in method section.

For the experiment, we hence select a cluttered scene with different objects: an
oil bottle, a tissue  box, a blue funnel, a drill, a cracker  box, a tomato soup
can, and a  spam can.  Including the background class  this defines an $8$-class
segmentation problem. Many  partial occlusions happen in this scene,  and as for
the first experiment, simply warping the  annotations is not sufficient. We show
that S4-Net has the ability of improving on such regions.

\begin{figure*}
\centering
\begin{subfigure}{.3\linewidth}
	\begin{tikzpicture}
	\node[] (simg) at (0.0, 0.0) {\includegraphics[trim=50 20 30 40, clip, width=1.\linewidth]{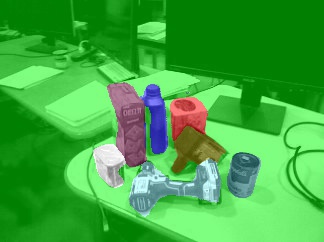}};
	
	\draw[color=black, line width=1mm] (-0.7,0) ellipse (3mm and 9mm);
	\end{tikzpicture}
\end{subfigure}
\begin{subfigure}{.3\linewidth}
	\includegraphics[trim=50 20 30 40, clip, width=1.\linewidth]{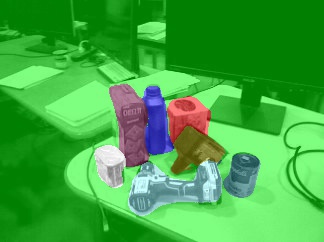}
\end{subfigure}
\begin{subfigure}{.3\linewidth}
	\includegraphics[trim=50 20 30 40, clip, width=1.\linewidth]{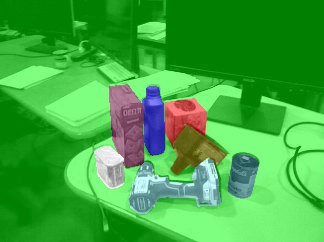}
\end{subfigure}
\begin{subfigure}{.3\linewidth}
	\begin{tikzpicture}
	\node[] (simg) at (0.0, 0.0) {\includegraphics[trim=50 20 30 40, clip, width=1.\linewidth]{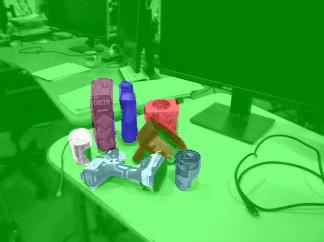}};
	
	\draw[color=black, line width=1mm] (-0.1,-0.1) ellipse (6mm and 8mm);
	\end{tikzpicture}
	
\end{subfigure}
\begin{subfigure}{.3\linewidth}
	\includegraphics[trim=50 20 30 40, clip, width=1.\linewidth]{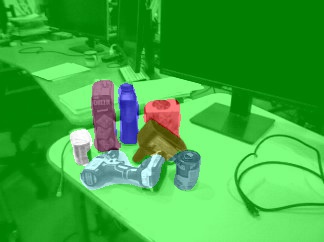}
\end{subfigure}
\begin{subfigure}{.3\linewidth}
	\includegraphics[trim=50 20 30 40, clip, width=1.\linewidth]{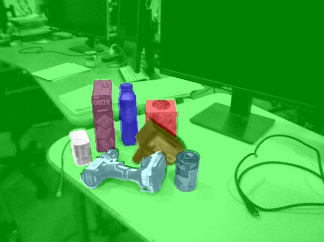}
\end{subfigure}
\begin{subfigure}{.3\linewidth}
	\begin{tikzpicture}
	\node[] (simg) at (0.0, 0.0) {\includegraphics[trim=50 20 30 40, clip, width=1.\linewidth]{figures/evaluation/object_cluter/comparison/sup_seg_00546}};
	
	\draw[color=black, line width=1mm] (-0.8,-0.4) ellipse (6mm and 8mm);
	\draw[color=black, line width=1mm] (-1.6,0.8) ellipse (4mm and 4mm);
	\end{tikzpicture}
	
	\caption{Supervised with annotation warping}
\end{subfigure}
\begin{subfigure}{.3\linewidth}
	\begin{tikzpicture}
	\node[] (simg) at (0.0, 0.0) {\includegraphics[trim=50 20 30 40, clip, width=1.\linewidth]{figures/evaluation/object_cluter/comparison/ss_seg_00546}};
	
	\draw[color=red, line width=1mm] (0.4,0.4) ellipse (10mm and 10mm);
	\end{tikzpicture}
	\caption{S4-Net}
\end{subfigure}
\begin{subfigure}{.3\linewidth}
	\includegraphics[trim=50 20 30 40, clip, width=1.\linewidth]{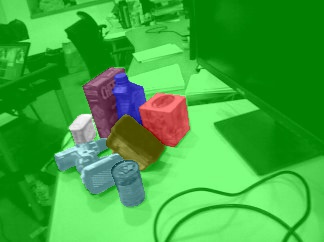}
		\caption{Ground Truth Annotation}
\end{subfigure}
		

\caption{Qualitative results on the segmentation problem from FusionLabel. Annotation warping on its own is not enough to learn a whole scene from a single manual annotation.  S4-NET improves performance on partially occluded objects up to some extent. Mistakes made by the supervised approach, shown with black ellipses, are corrected by S4-Net because of the influence of geometric constraint. However, S4-Net fails to deal with the regions shown with red ellipses that consistently output the wrong class. }
\label{fig:clutter_comp}
\end{figure*}

Figure~\ref{fig:clutter_comp}  shows   some  of  the   resulting  segmentations.
Supervised  learning with  annotation warping  performs well.   Even though  the
annotations can  not be warped  on every surface when  two objects are  close to
each other, supervised learning is still able to learn segmentations for some of
such  regions because  of the  symmetry in  the textures.   However, perspective
differences  limit this  generalization. The  geometric constraint  exploited by
S4-Net improves the  results on those regions by comparing  such features warped
under different perspectives. Finally, we performed a quantitative evaluation of the experiment. We calculate the IoU scores for both the supervised  approach and S4-Net. Supervised approach with annotation warping achieves score of $86.7$. S4-Net is able to reach even better results by achieving score of $88.4$ and therefore outperforming the supervised approach.

\newpage
\section{Conclusion}

We presented S4-Net, a semi-supervised semantic segmentation learning method with geometric constraints. We showed that  semi-supervised learning with geometric constraints is a powerful  framework to efficiently
learn semantic segmentation when only a fraction of data is annotated. The current  drawback of our implementation is that
it  requires  depth  data  for  the   training  images.   As  mentioned  in  the
introduction, an  option is to  rely on automated  dense SLAM systems  to obtain
such   data  from   color  images.   Such  systems   are  also   in  very   fast
development~\cite{Godard17,ZhouBSL17,Bloesch18},  and a  very exciting  possible
extension of  our work is to  simultaneously learn to recover  both geometry and
semantic information from simple video sequences.

\section*{Acknowledgment}

This work was supported by the Christian Doppler Laboratory for Semantic 3D Computer Vision, funded in part by Qualcomm Inc.

{\small
\bibliographystyle{ieee}
\bibliography{egbib}
}

\end{document}